\pdfoutput=1
\documentclass[conference]{IEEEtran}
\IEEEoverridecommandlockouts
\makeatletter
\newcommand*\titleheader[1]{\gdef\@titleheader{#1}}
\AtBeginDocument{%                                                               
  \let\st@red@title\@title
  \def\@title{%                                                                 
    \bgroup\normalfont\normalsize\centering\@titleheader\par\egroup
    \vskip1ex\st@red@title}
}
\makeatother

\usepackage{cite}
\usepackage{amsmath,amssymb,amsfonts}
\usepackage{algorithmic}
\usepackage{array}
\usepackage{graphicx}
\usepackage{textcomp}
\usepackage[table]{xcolor}
\usepackage{multirow}
\usepackage{subcaption}

\def\BibTeX{{\rm B\kern-.05em{\sc i\kern-.025em b}\kern-.08em
    T\kern-.1667em\lower.7ex\hbox{E}\kern-.125emX}}
\usepackage{listings}

\definecolor{codegreen}{rgb}{0,0.6,0}
\definecolor{codegray}{rgb}{0.5,0.5,0.5}
\definecolor{codepurple}{rgb}{0.58,0,0.82}
\definecolor{backcolour}{rgb}{0.95,0.95,0.92}

\lstdefinestyle{mystyle}{
    backgroundcolor=\color{backcolour},   
    commentstyle=\color{codegreen},
    keywordstyle = {\color{magenta}},
    keywordstyle = [2]{\color{lime}},
    keywordstyle = [3]{\color{yellow}},
    keywordstyle = [4]{\color{teal}},
    numberstyle=\tiny\color{codegray},
    stringstyle=\color{codepurple},
    basicstyle=\ttfamily\footnotesize,
    breakatwhitespace=false,         
    breaklines=true,                 
    captionpos=b,                    
    keepspaces=true,                 
    numbers=left,                    
    numbersep=5pt,                  
    showspaces=false,                
    showstringspaces=false,
    showtabs=false,                  
    tabsize=2
}

\title{MicroViT: A Vision Transformer with Low Complexity Self Attention for Edge Device \\
}
\author{\IEEEauthorblockN{Novendra Setyawan\textsuperscript{1,3}, Chi-Chia Sun\textsuperscript{*,2}, Mao-Hsiu Hsu\textsuperscript{1}, Wen-Kai Kuo\textsuperscript{1}, Jun-Wei Hsieh\textsuperscript{4}}
\IEEEauthorblockA{\textit{\textsuperscript{1}Department of Electro-Optics Engineering, National Formosa University, Taiwan}\\
\textit{\textsuperscript{2}Department of Electrical Engineering, National Taipei University, Taiwan} \\
\textit{\textsuperscript{3}Department of Electrical Engineering, University of Muhammadiyah Malang, Indonesia} \\
\textit{\textsuperscript{4}College of Artificial Intelligence and Green Energy, National Yang Ming Chiao Tung University, Taiwan} \\
\textit{chichiasun@gm.ntpu.edu.tw\textsuperscript{*}}
} 
}

\titleheader{\vspace{-20pt}To appear at the 2025 IEEE International Symposium on Circuits and Systems (ISCAS), May 25-28 2025, London, UK.}

\begin{document}

\maketitle

\begin{abstract}

The Vision Transformer (ViT) has demonstrated state-of-the-art performance in various computer vision tasks, but its high computational demands make it impractical for edge devices with limited resources. This paper presents MicroViT, a lightweight Vision Transformer architecture optimized for edge devices by significantly reducing computational complexity while maintaining high accuracy. The core of MicroViT is the Efficient Single Head Attention (ESHA) mechanism, which utilizes group convolution to reduce feature redundancy and processes only a fraction of the channels, thus lowering the burden of the self-attention mechanism. MicroViT is designed using a multi-stage MetaFormer architecture, stacking multiple MicroViT encoders to enhance efficiency and performance. Comprehensive experiments on the ImageNet-1K and COCO datasets demonstrate that MicroViT achieves competitive accuracy while significantly improving  $3.6 \times$ faster inference speed and reducing energy consumption with 40\% higher efficiency than the MobileViT series, making it suitable for deployment in resource-constrained environments such as mobile and edge devices.

\end{abstract}

\begin{IEEEkeywords}
Classification, Self Attention, Vision Transformer, Edge Device.
\end{IEEEkeywords}

\section{Introduction}
In recent years, Transformers have gained significant attention and demonstrated remarkable achievements in computer vision. A notable development in this field was the introduction of the Vision Transformer (ViT) \cite{dosovitskiy2020image}, which utilizes pure Transformers for image classification tasks. Following ViT, several models have been proposed to improve performance, achieving promising results in a range of vision tasks, including image classification, object detection, and segmentation \cite{setyawan2024multi, liang2024swin, gao2023metformer, yu2024spikingvit, hsu2024inpainting}.

Despite the strong performance of the vanilla Vision Transformer (ViT) \cite{dosovitskiy2020image}, it requires between 85 million and 632 million parameters to handle ImageNet classification tasks. Its forward propagation demands substantial computational resources, resulting in slow inference speeds and making it unsuitable for many specific applications. Applying Transformer models in low-cost settings, such as mobile and edge devices, is particularly challenging due to constraints on memory, processing power, and battery life \cite{setyawan2024fpga}. Therefore, our work focuses on constructing a lightweight and efficient deep learning model that reduces computational requirements and power consumption, while ensuring fast inference and high performance on edge devices.

Several studies have attempted to lower the computational complexity of Vision Transformers by integrating them with Convolutional Neural Networks (CNNs) \cite{mehta2021mobilevit, wu2021cvt}. For example, MobileViT uses an inverted bottleneck convolution block from MobileNetV2 \cite{sandler2018mobilenetv2} in the early stages to reduce the computational complexity of ViT. Numerous studies \cite{wang2021pyramid, mehta2022separable, maaz2022edgenext} have identified that the self-attention (SA) mechanism, particularly in the spatial feature mixer of Transformers, is the most computationally demanding component. Pyramid Vision Transformer (PVT) reduces the quadratic complexity ($\mathcal{O}(n^2)$) of SA by applying Spatial Reduction Attention (SRA) to shorten the token length. MobileViTv2 proposes Separable Linear Attention \cite{mehta2022separable} to alleviate the burden of SA, while EdgeNeXt \cite{maaz2022edgenext} employs transposed SA to tackle the complexity challenges for edge device implementation. Another approach, SHViT, suggests that Multi-Head Self-Attention (MHSA) can suffer from feature redundancy across heads, and introduces single-head attention that processes only a quarter of the image tokens or features. However, most of these methods are developed separately, without considering power consumption in limited-resource environments like edge computing devices. 

This paper introduces MicroViT, a novel Vision Transformer model optimized for edge device deployment. The proposed architecture significantly reduces computational complexity and power consumption by utilizing Efficient Single Head Attention (ESHA), which minimizes feature redundancy through group convolutions and single-head SA, processing only a quarter of the overall channels. Built upon the MetaFormer architecture, MicroViT ensures fast inference and low power usage while maintaining high accuracy while evaluated on edge devices, making it ideal for energy-constrained environments.
\begin{figure*}
    \centering
    \includegraphics[width=15cm]{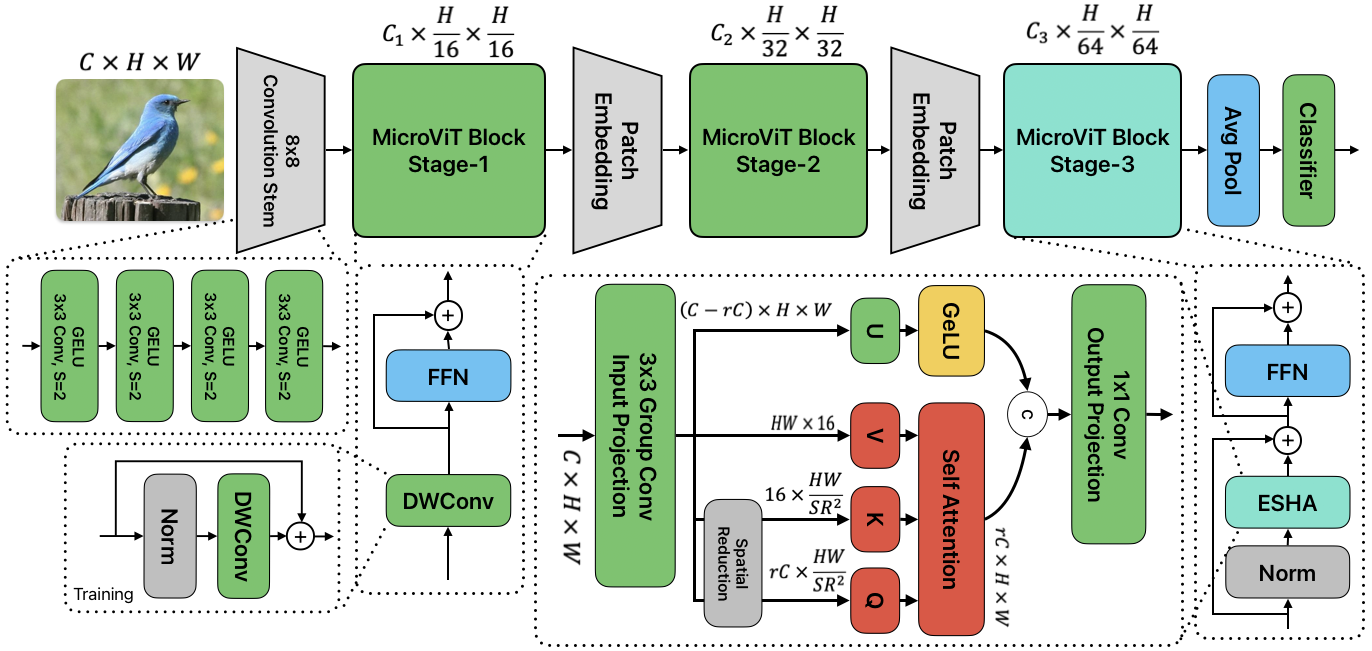}
    \caption{\textbf{MicroViT} conducted with 4 stage pyramid feature map MetaFormer architecture. The first two stage use Depth-Wise Convolution (DWConv) and the last two stage use Low Resolution Single Head Attention as spatial feature mixer.}
    \label{fig:MicroViT-Arch}
\end{figure*}
\section{Related Work}

Several lightweight models have been created to address the computational challenges of Vision Transformers (ViTs) in resource-constrained settings. Notable examples include MobileViT \cite{mehta2021mobilevit}, FastViT \cite{vasu2023fastvit}, and EfficientFormerV2 \cite{li2023rethinking}. These models integrate convolutional units from MobileNetV2 \cite{sandler2018mobilenetv2} or MobileOne \cite{vasu2023mobileone} with Transformer architectures. They employ spatially efficient convolutions in the initial stages to minimize computational cost, while Transformer layers handle long-range dependencies, achieving strong image classification performance on edge and mobile devices.

Efficient self-attention (SA) mechanisms are crucial for ViT scalability. MobileViTv2 \cite{mehta2022separable} employs Separable Linear Attention to separate spatial and channel interactions, reducing computational load, and enhancing edge computing efficiency. Similarly, EdgeNeXt \cite{maaz2022edgenext} applies transposed SA to boost performance on low-power devices, preserving accuracy. Additionally, SHViT (Single-Head Vision Transformer) \cite{yun2024shvit} streamlines attention by substituting the traditional MHSA with a single head, cutting redundant feature extraction and computational complexity while maintaining accuracy.

However, there remains a gap in the literature concerning holistic approaches that simultaneously address computational efficiency, feature redundancy, and power consumption. Our proposed MicroViT aims to bridge this gap by offering a lightweight Vision Transformer architecture optimized for both
performance and power efficiency in edge device deployments.
\section{Proposed Method}
The main idea of a novel MicroViT architecture for an efficient vision model is reducing the computational complexity and redundancy for processing the image features. The MicroViT model incorporates the Efficient Single Head Attention (ESHA) technique, which could generate low-redundancy feature maps with low complexity SA.

\subsection{Efficient Single Head Attention (ESHA)} 
ESHA combines local and global spatial operations within a single block for efficient token information extraction. Unlike the original SA, which employs linear or PWConv to form queries, keys, and values, ESHA uses kernel-based group convolution for local information acquisition. In SA, tokens undergo scaled dot operations to retrieve the global feature context. Specifically, if $X_i$ is the feature map from the patch embedding block, ESHA is projected into $Q\in \mathbb{R}^{H\times W\times C_{q}}$, $K\in \mathbb{R}^{H\times W\times C_{k}}$, $V\in \mathbb{R}^{H\times W\times C_{v}}$, and $U\in \mathbb{R}^{H\times W\times C_{u}}$ as query, key, value, and unaltered feature. In Equation \ref{eq:in_proj}, $W_{ip}$ represents a $3 \times 3$ convolution kernel with 32 groups to reduce computation.
\begin{align}
    Q, K, V, U &= Split(W_{ip}*X_i), 
    \label{eq:in_proj}
\end{align} 
The projected input will be spliced that corresponds to the channel as described in equation \ref{eq:in_proj}. The query and key dimensions $C_q$ and $C_k$ are set to 16 as the maximum. The channel numbers of value and un-touch $C_v$ and $C_u$ follow the channel ratio $r$ with $rC$ for v and $2(C-C_a)$ for un-touch channel as illustrated in Figure \ref{fig:MicroViT-Arch}. The optimal channel number ratio is set $r=0.215$ that has been discussed in \cite{yun2024shvit} in single head attention.
The global attention score $A\in \mathbb{R}^{H\times W\times rC_{i}}$ will be processed with the scaled dot product operation in the following Equation \ref{eq:att_dw}.
\begin{align}
    A &= V\cdot Softmax(Q^T\cdot K) 
    \label{eq:att_dw}
\end{align} 
For fast computation, the spatial or token length will be reduced in $K\in \mathbb{R}^{\frac{H}{SR}\times \frac{W}{SR}\times C_{k}}$ and $V\in \mathbb{R}^{\frac{H}{SR}\times \frac{W}{SR}\times C_{v}}$ using DWConv with $SR$ ratio. It will reduce the computation of the dot product with $SR^2$ reduction. Next, the convolution projection will ensure efficient propagation of the attention features with un-touch from the previous convolution feature as detailed in equation \ref{eq:esha_mixer}.
\begin{align}
    X_i' &= W_{op}*Cat(A,\sigma{U})
    \label{eq:esha_mixer}
\end{align} 
The $\sigma$ denotes the activation function, such as $GELU(.)$ as the commonly used. The $W_{op}$ is the weight of $1 \time 1$ convolution operation that will connect feature across the channel. 

\begin{table}[ht]
\begin{center}
\caption{All Variant MicroViT Model configurations. \#Blocks denotes number of blocks. ESHA Dim means the number channel of feature map.}
\begin{tabular}{ccccccc}
\hline
\multicolumn{1}{c|}{\multirow{2}{*}{Stage}} & \multicolumn{1}{c|}{\multirow{2}{*}{Size}} & \multicolumn{1}{c|}{\multirow{2}{*}{Layer}} & \multicolumn{3}{c}{MicroViT}  \\ 
\cline{4-6} 
\multicolumn{1}{l|}{}  & \multicolumn{1}{l|}{} & \multicolumn{1}{l|}{} & \multicolumn{1}{c|}{S1} & \multicolumn{1}{c|}{S2} & \multicolumn{1}{c}{S3}  \\ 
\hline
\multicolumn{1}{c|}{\multirow{3}{*}{1}} & \multicolumn{1}{l|}{\multirow{3}{*}{$\frac{H}{16}\times\frac{W}{16}$}} & \multicolumn{1}{l|}{\multirow{1}{*}{Stem}} & \multicolumn{3}{c}{[3x3 Conv S=2, GeLU]$\times$4} \\
\cline{3-6} 
\multicolumn{1}{l|}{} & \multicolumn{1}{l|}{} & \multicolumn{1}{l|}{DWConv ($C$)}  & \multicolumn{1}{c|}{ 128 } & \multicolumn{1}{c|}{ 128 } & \multicolumn{1}{c}{192}    \\
\cline{3-6} 
\multicolumn{1}{l|}{} & \multicolumn{1}{l|}{} & \multicolumn{1}{l|}{\#Blocks} & \multicolumn{1}{c|}{2}  & \multicolumn{1}{c|}{2}  & \multicolumn{1}{c}{3} \\ 
\hline
\multicolumn{1}{c|}{\multirow{3}{*}{2}} & \multicolumn{1}{l|}{\multirow{3}{*}{$\frac{H}{32}\times\frac{W}{32}$}} & \multicolumn{1}{l|}{\begin{tabular}[c]{@{}l@{}}Patch Embed\end{tabular}} & \multicolumn{3}{c}{3x3, Stride 2} \\
\cline{3-6} 
\multicolumn{1}{l|}{} & \multicolumn{1}{l|}{} & \multicolumn{1}{l|}{DWConv ($C$)} & \multicolumn{1}{c|}{256} & \multicolumn{1}{c|}{320} & \multicolumn{1}{c}{384}  \\ 
\cline{3-6}
\multicolumn{1}{l|}{} & \multicolumn{1}{l|}{} & \multicolumn{1}{l|}{\#Blocks} & \multicolumn{1}{c|}{5} & \multicolumn{1}{c|}{7} & \multicolumn{1}{c}{6} \\ 
\hline
\multicolumn{1}{c|}{\multirow{7}{*}{3}} & \multicolumn{1}{l|}{\multirow{7}{*}{$\frac{H}{64}\times\frac{W}{64}$}}  & \multicolumn{1}{l|}{\begin{tabular}[c]{@{}l@{}}Patch Size\end{tabular}} & \multicolumn{3}{c}{3x3, Stride 2} \\ 
\cline{3-6} 
\multicolumn{1}{l|}{} & \multicolumn{1}{l|}{} & \multicolumn{1}{l|}{ESHA Dim} & \multicolumn{1}{c|}{320}  & \multicolumn{1}{c|}{448} & \multicolumn{1}{c}{512} \\ 
\multicolumn{1}{l|}{}  & \multicolumn{1}{l|}{}  & \multicolumn{1}{l|}{QK ($C_q, C_k$)} & \multicolumn{1}{c|}{16, 16}  & \multicolumn{1}{c|}{16, 16} &  \multicolumn{1}{c}{16, 16} \\ 
\multicolumn{1}{l|}{} & \multicolumn{1}{l|}{} & \multicolumn{1}{l|}{ratio $r$}  & \multicolumn{1}{c|}{$1/4$} & \multicolumn{1}{c|}{$1/4$}  & \multicolumn{1}{c}{$1/4$}  \\
\multicolumn{1}{l|}{}  & \multicolumn{1}{l|}{} & \multicolumn{1}{l|}{Group ($g$)} & \multicolumn{1}{c|}{32}  & \multicolumn{1}{c|}{32} &  \multicolumn{1}{c}{32} \\ 
\multicolumn{1}{l|}{}  & \multicolumn{1}{l|}{}  & \multicolumn{1}{l|}{SR} & \multicolumn{1}{c|}{2}  & \multicolumn{1}{c|}{2} &  \multicolumn{1}{c}{1} \\ 
\cline{3-6} 
\multicolumn{1}{l|}{} & \multicolumn{1}{l|}{} & \multicolumn{1}{l|}{\#Blocks}  & \multicolumn{1}{c|}{5} & \multicolumn{1}{c|}{5} & \multicolumn{1}{c}{6} \\ 
\hline
\multicolumn{3}{c|}{Classifier Head} & \multicolumn{3}{c}{Avg Pool, FC }\\ \hline
\end{tabular}
\label{tab:arch_variant}
\end{center}
\end{table}

\subsection{Overall Architechture}
MicroViT employed a three-stage pyramid structure, starting with a $16\times 16$ stem comprising four $3\times3$ convolutions to shrink features by a factor of 16. Adopting the MetaFormer \cite{yu2022metaformer} framework, it utilized two residual blocks for spatial mixing, followed by a residual Feed Forward Network (FFN) for channel mixing, as explained in Equation \ref{eq:encoder}.
\begin{align}
    X'_i &= X_i+\lambda_i \odot SpatialMixer(X_i)\\
    X''_i &= X'_i+\lambda_i \odot FFN(X'_i)
    \label{eq:encoder}
\end{align} 
where $X_i', X_i'' \in \mathbb{R}^{C_i \times H\times W } $. The FFN block contains a sequence of point-wise convolutions as a linear operation. It encompasses a singular activation function, which may be mathematically represented by the subsequent equation.
\begin{equation}
    FFN(X'_i)=(\sigma(Norm(X'_i)*W_{fc1}))*W_{fc2},
\end{equation}
 The $W_{fc1} \in \mathbb{R}^{C_i\times \alpha C_i}$ and $W_{fc2} \in \mathbb{R}^{\alpha C_i\times C_i}$ are learnable weights with $\alpha$ expansion ratio with a default of 2. The $\sigma$ denotes the $GELU(.)$ activation function.

The MicroViT model employs a sequence of separable convolutions and a residual FFN for patch embedding, offering reduction rates of 32 in stage-2 and 64 in stage-3, and uses a $3 \times 3$ patch embedding. In the early stages, DW convolution acts as spatial mixers to fulfill higher memory demands. The final stage utilizes the Efficient Single Head Attention (ESHA) mechanism as outlined in Table \ref{tab:arch_variant}. Batch Normalization (BN) is used to better integrate with adjacent convolutional layers and reduce reshaping, thus enhancing inference speed. The architecture uses global average pooling followed by a fully connected layer for feature extraction and classification.

\section{Result}
For the evaluation of MicroViT, the ImageNet-1K dataset \cite{russakovsky2015imagenet} comprising 1.28 million training images and 50,000 validation images over 1,000 categories was employed. Following the DeiT training method \cite{touvron2021training}, models were trained for 300 epochs at a 224×224 resolution with an initial learning rate of 0.004, utilizing various data augmentations. The AdamW optimizer \cite{loshchilov2017decoupled} was used with a batch size of 512 across three A6000 GPUs.

We assessed model throughput in various computation environments, including a GPU (RTX-3090), a CPU (Intel i5-13500), and specifically the Jetson Orin Nano edge device. For throughput, the GPU and CPU had a batch size of 256, whereas the edge device used a batch size of 64 with ONNX Runtime. To enhance performance during inference, BN layers were fused with adjacent layers when possible. On the Jetson Orin Nano, we also examined power and energy usage during latency tests with 1000 images at a consistent resolution.

We further evaluate MicroViT on object detection on the COCO dataset \cite{lin2014microsoft} utilizing RetinaNet \cite{ross2017focal} and conduct training for 12 epochs (1$\times$ schedule), adhering to the configuration used by \cite{liu2023efficientvit} in mmdetection \cite{mmdetection}. In the object detection experiments, we employ AdamW \cite{loshchilov2017decoupled} with a batch size of 16, a learning rate of $1\times10^{-3}$, and a weight decay rate of 0.025.

\begin{table}[!ht]
\centering
\caption{Comparison of All MicroViT Variant with SOTA on ImageNet-1K Dataset. Res, Par and FLPs denotes as input resolution, parameters and Floating Operation. GPU and CPU denotes a inference throughput (img/s) in device respectively.}
\begin{tabular}{ m{2.6cm}|c|c|c|c|>{\centering}m{0.5cm}|c }
\hline
Model   & Res  & Par & FLPs &  GPU  & CPU &   Top-1      \\ \hline
% Fasternet-T1\cite{liu2022convnet}              & 224 & 7.6  & 0.85 & \textbf{7037}  & 149.3  &   & 76.2      \\
MobileNetV2-1.0\cite{sandler2018mobilenetv2}& 224 & 3.5 & 0.314 & 4527 & 82 & 72.0   \\
MobileViT-XXS\cite{mehta2021mobilevit}  & 256 & 1.3 & 0.261 & 3218 & 99  & 69.0   \\
MobileViTV2-0.5\cite{mehta2022separable}  & 256 & 1.4 & 0.480 & 3885 & 68  & 70.2   \\
EdgeNeXt-XXS\cite{maaz2022edgenext}  & 256 & 1.3 & 0.261 & 3975 & 245  & 71.2   \\
Fasternet-T0\cite{chen2023run}  & 224 & 3.9 & 0.340 & 11775 & 311 & 71.9   \\
SHViT-S1\cite{yun2024shvit} & 224  & 6.3 & 0.241 & 15280 & 475 & 72.8 \\
\rowcolor{gray!30}
\textbf{MicroViT-S1}                    & 224  & 6.4 & 0.231 & 17466 & 552 & 72.6 \\ \hline 
EFormerV2-S0\cite{li2023rethinking}& 224  & 3.6 & 0.407 & 1191 & 91 & 73.7 \\ 
EdgeNeXt-XS\cite{maaz2022edgenext}   & 256 & 2.3 & 0.536 & 2935 & 139  & 75.0   \\
EfficientViT-M4\cite{liu2023efficientvit} & 224 & 8.8 & 0.303 & 10093 & 379 & 74.3   \\
MobileViT-XS\cite{mehta2021mobilevit}   & 256 & 2.3 & 0.935 & 1740 & 43 & 74.8   \\
MobileNetV3-L\cite{howard2019searching}& 224 & 3.5 & 0.314 & 4527 & 82 & 75.2  \\
SHViT-S2\cite{yun2024shvit}            & 224  & 11.5 & 0.366 & 12007 & 367 & 75.2 \\
\rowcolor{gray!30}
\textbf{MicroViT-S2}                    & 224 & 10.0 & 0.345 & 14154 & 435 & 74.6 \\ \hline
FastViT-T8\cite{vasu2023fastvit}   & 256 & 4.0 & 0.687 & 3719 & 83 & 76.2   \\
Fasternet-T1\cite{chen2023run}  & 224 & 7.6 & 0.851 & 7151 & 130 & 76.2   \\
EfficientViT-M5\cite{liu2023efficientvit}& 224 & 12.5 & 0.526 & 6807 & 233 & 77.1   \\
SHViT-S3\cite{yun2024shvit}            & 224  & 14.1 & 0.601 & 8180 & 224 & 77.8 \\
% \hline 
\rowcolor{gray!30}
\textbf{MicroViT-S3}                   & 224 & 16.7 & 0.580 & 9288 & 232 & 77.1 \\ \hline 
    \end{tabular}
    \label{tab:imgnet-result}
\end{table}
\subsection{ImageNet-1K Classification Result}
Table \ref{tab:imgnet-result} presents a comparison of various MicroViT variants with state-of-the-art (SOTA) models on the ImageNet-1K dataset. The evaluation focuses on models' computational efficiency and accuracy, highlighting the trade-offs between resource consumption and performance. 

MicroViT-S1 demonstrated superior performance compared to traditional CNN models, surpassing MobileNetV2-1.0\cite{sandler2018mobilenetv2} and Fasternet-T0\cite{mehta2022separable}, with a $3.6 \times$ faster in GPU and $6.7 \times$ in CPU throughput, while maintaining an accuracy advantage of 0.8 over MobileNetV2-1.0. Additionally, MicroViT-S2 outperformed mobile transformers like EfficientFormerV2-S0\cite{li2023rethinking} and EfficientViT-M4\cite{liu2023efficientvit}, achieving $0.3\%$ better accuracy with similar efficiency metrics. Across the MicroViT models, CPU throughput is robust, notably with MicroViT-S1 achieving 552 img/s, which is $8 \times$ faster than several EfficientViT variants, illustrating MicroViT's adaptability to both high-end GPU and CPU settings.

Table \ref{tab:edge-result} presents the performance of MicroViT variants against various SOTA models on the Edge device using ONNX. MicroViT-S1's throughput reaches 773 img/s, efficiently managing large image volumes on the Jetson Orin Nano. This surpasses several SOTA models like MobileViT-XS\cite{mehta2021mobilevit} and EfficientFormer-V2-S0\cite{li2023rethinking}, making MicroViT-S1 optimal for rapid image processing applications. Furthermore, it has a 9.1 ms latency, outperforming MobileNetV2-1.0\cite{sandler2018mobilenetv2} and EdgeNeXt-XS\cite{maaz2022edgenext}, supporting real-time use. It consumes 2147 Joules, achieving high energy efficiency with $\eta=3.7$. Likewise, MicroViT-S2 and MicroViT-S3 balance throughput and energy use, maintaining accuracy, thus ideal for resource-constrained edge devices with superior power efficiency over other lightweight vision transformers.
\begin{table}[!ht]
\centering
\caption{Comparison of All MicroViT Variant and SOTA on ImageNet-1K Dataset with NVIDIA Jetson Orin Nano Edge Device using ONNX format.}
\begin{tabular}{ m{2.6cm}|>{\centering}m{0.5cm}|>{\centering}m{0.6cm}|>{\centering}m{0.9cm}|>{\centering}m{0.8cm}|c }
\hline
\multirow{2}{*}{Model} & Thg & Lat & Avg Pow & Energy  & $\eta$ \\ 
         & img/s & (ms) & (W) & (Joule) & \% / J \\ \hline
MobileNetV2-1.0\cite{sandler2018mobilenetv2} & 234 & 6.7 & 3549 & 23.9 & 3.01 \\
MobileViT-XXS\cite{mehta2021mobilevit}  & 184 & 9.6 & 3428 & 32.4 & 2.13   \\
MobileViTV2-0.5\cite{mehta2022separable}  & 208 & 10.9 & 2887 & 31.6 & 2.22 \\
EdgeNeXt-XXS\cite{maaz2022edgenext}     & 257 & 8.1 & 2805 & 22.6 &  3.15  \\
Fasternet-T0\cite{chen2023run}   & 675 & 8.4 & 2419 & 20.4 &  3.52 \\
SHViT-S1 \cite{yun2024shvit}       & 813  & 12.6 & 2069 & 26.0 & 2.80  \\
\rowcolor{gray!30}
\textbf{MicroViT-S1}                & 773  & 9.1 & 2147 & 19.6 & 3.7 \\ \hline 
EFormerV2-S0\cite{li2023rethinking}  & 257  & 10.9 & 2847 & 31.1 & 2.37   \\ 
EdgeNeXt-XS\cite{maaz2022edgenext}      & 168 & 11.1 & 3031 & 33.6 &  2.32   \\
EfficientViT-M4\cite{liu2023efficientvit} & 587 & 18.0 & 2019 & 36.3 &  2.05   \\
MobileViT-XS\cite{mehta2021mobilevit}   & 96.6 & 14.3 & 3730 & 53.3 & 1.40    \\
MobileNetV3-L\cite{howard2019searching} & 310 & 8.3 & 2743 & 22.6 &  3.33 \\
SHViT-S2 \cite{yun2024shvit}            & 598  & 12.7 & 2306 & 29.4 & 2.56  \\
\rowcolor{gray!30}
\textbf{MicroViT-S2}                      & 567 & 9.9 & 2481 & 24.3 & 3.07 \\ \hline
FastViT-T8\cite{vasu2023fastvit}          & 176 & 8.6 & 3622 & 30.8 &  2.47  \\
Fasternet-T1\cite{chen2023run}     & 421 & 8.8 & 2860 & 25.3 &  3.01  \\
EfficientViT-M5\cite{liu2023efficientvit} & 409 & 21.0 & 2180 & 45.9 & 1.68    \\
SHViT-S3\cite{yun2024shvit}               & 425  & 14.6 & 2500 & 36.6 & 2.13  \\
% \hline 
\rowcolor{gray!30}
\textbf{MicroViT-S3}                    & 398 & 10.9 & 2609 & 28.6 & 2.69  \\ \hline 
    \end{tabular}
    \label{tab:edge-result}
\end{table}

\subsection{Object Detection Result}
We compare MicroViT-3 with efficient models \cite{sandler2018mobilenetv2, howard2019searching, liu2023efficientvit} on the COCO \cite{lin2014microsoft} object detection task, and present the results in Table \ref{tab:obj} Specifically, MicroViT-3 surpasses MobileNetV2 \cite{sandler2018mobilenetv2} by
7.7\% AP with comparable Flops. Compared to the EfficientViT-M4, our MicroViT-3 uses 46.8\% fewer Flops while achieving 3.3\% higher AP, demonstrating its capacity and generalization ability in different vision tasks.

\begin{table}[!ht]
\centering
\caption{Object detection on COCO val2017 with RetinaNet. $AP^b$ denote bounding box average precision. The FLOPs (G) are measured at resolution 1280 $\times$ 800.
}
\begin{tabular}{ m{2.4cm}|>{\centering}m{0.8cm}|>{\centering}m{0.8cm}|>{\centering}m{0.8cm}|>{\centering}m{0.8cm}|c}
\hline
Backbone & $AP^{b}$  & $AP^{b}_{50}$ & $AP^{b}_{75}$ & Par & FLPs  \\ \hline
MobileNetV2\cite{sandler2018mobilenetv2} & 28.3& 46.7& 29.3& 3.4& 300 \\
MobileNetV3\cite{howard2019searching} & 29.9& 49.3& 30.8& 5.4 & 217   \\ 
EfficientViT-M4\cite{liu2023efficientvit}&32.7 &52.2 &34.1 & 8.8 & 299  \\
\rowcolor{gray!30}
\textbf{MicroViT-S3} & 36.0 & 56.6 & 38.2 & 26.7 & 159    \\ \hline

\end{tabular}
\label{tab:obj}
\end{table}

\subsection{Ablation Study}

Table \ref{tab:abl-result} assesses the effects of three ablations on the MicroViT-S2 model as the baseline. The first ablation, low resolution SA with SR=2, slightly increases the parameter count from 11.0M to 11.1M, with GPU throughput remaining nearly unchanged from 12009 to 12029 images per second. However, it increases latency from 9.9 ms to 10.7 ms and decreases Top-1 accuracy to 72.5\%, resulting in a minor drop in efficiency ($\eta$) from 3.0 to 2.8. This indicates a small trade-off between accuracy and computational cost due to the architectural modification.

The Next ablation, without group convolutions, increases the model's complexity significantly to 20.5M parameters, with higher GFLOPs. This results in lower throughput and efficiency ($\eta=1.8$) but achieves the highest accuracy. However, this variant consumes more energy, making it ideal for use cases where accuracy is prioritized over efficiency.

The ablation study shows that even though spatial reduction can increase throughput, the latency is increased, resulting in an efficiency drop. Group convolution successfully increases the efficiency in ESHA compared to other attention models, such as vanilla attention in MobileViT, maintaining lower complexity and energy usage.

\begin{table}[!ht]
\centering
\caption{Ablation study of MicroViT  on ImageNet-1K Dataset. The Baseline model is MicroViT-S2.}
\begin{tabular}{ m{1.6cm}|c|c|c }
\hline
\textbf{Ablation} & Baseline & Low Res Attn & W/O Group  \\ \hline
\textbf{Param}      & 11.0  & 11.1  & 20.5   \\  
\textbf{GFLOPs}     & 0.343 & 0.344 & 0.4977 \\ 
\textbf{GPU}        & 14154 & 14179 & 13511  \\
\textbf{Throughput} & 567   & 570   & 503    \\
\textbf{Latency}    & 9.9   & 10.7  & 12.5   \\
\textbf{Avg Pow}    & 2481  & 2512  & 3288   \\
\textbf{Energy}     & 24.29  & 25.8  & 41.2   \\
\textbf{Top-1}      & 72.7  & 72.5  & 73.2   \\
\textbf{Efficiency ($\eta$)}& 3.07  & 2.8 & 1.8 \\
\hline

    \end{tabular}
    \label{tab:abl-result}
\end{table}

\section{Conclusion}
This paper introduces MicroViT, a novel lightweight Vision Transformer architecture optimized for edge devices, considering computational power and energy efficiency. By employing the Efficient Single Head Attention (ESHA) mechanism, MicroViT achieves a substantial reduction in computational complexity and energy consumption while maintaining competitive accuracy in vision tasks. Extensive experiments on the ImageNet-1K and COCO datasets demonstrate that MicroViT not only improves $3.6 \times$ throughput and inference speed but also surpasses several MobileViT models with 40\% efficiency and performance on edge devices. These results confirm that MicroViT is a promising solution for deploying Vision Transformers in resource-constrained environments. Future work will explore further optimizations and broader applications of this architecture in other edge computing tasks.

\bibliography{biblio}

\begin{thebibliography}{10}

\bibitem{dosovitskiy2020image}
A.~Dosovitskiy, ``An image is worth 16x16 words: Transformers for image recognition at scale,'' {\em arXiv preprint arXiv:2010.11929}, 2020.

\bibitem{setyawan2024multi}
N.~Setyawan, M.~N. Achmadiah, C.-C. Sun, and W.-K. Kuo, ``Multi-stage vision transformer for batik classification,'' in {\em 2024 International Electronics Symposium (IES)}, pp.~449--453, IEEE, 2024.

\bibitem{liang2024swin}
J.-A. Liang and J.-J. Ding, ``Swin transformer for pedestrian and occluded pedestrian detection,'' in {\em 2024 IEEE International Symposium on Circuits and Systems (ISCAS)}, pp.~1--5, IEEE, 2024.

\bibitem{gao2023metformer}
J.~Gao, K.-H. Yap, Y.~Wang, K.~Garg, and B.~S. Han, ``Metformer: A motion enhanced transformer for multiple object tracking,'' in {\em 2023 IEEE International Symposium on Circuits and Systems (ISCAS)}, pp.~1--5, IEEE, 2023.

\bibitem{yu2024spikingvit}
L.~Yu, H.~Chen, Z.~Wang, S.~Zhan, J.~Shao, Q.~Liu, and S.~Xu, ``Spikingvit: a multi-scale spiking vision transformer model for event-based object detection,'' {\em IEEE Transactions on Cognitive and Developmental Systems}, 2024.

\bibitem{hsu2024inpainting}
M.-H. Hsu, Y.-C. Hsu, and C.-T. Chiu, ``Inpainting diffusion synthetic and data augment with feature keypoints for tiny partial fingerprints,'' {\em IEEE Transactions on Biometrics, Behavior, and Identity Science}, 2024.

\bibitem{setyawan2024fpga}
N.~Setyawan, C.-C. Sun, W.-K. Kuo, and M.-H. Hsu, ``Fpga-based batik classification using quantization aware training of mobilenet and data-flow implementation,'' in {\em 2024 IEEE Asia Pacific Conference on Circuits and Systems (APCCAS)}, pp.~306--310, IEEE, 2024.

\bibitem{mehta2021mobilevit}
S.~Mehta and M.~Rastegari, ``Mobilevit: light-weight, general-purpose, and mobile-friendly vision transformer,'' {\em arXiv preprint arXiv:2110.02178}, 2021.

\bibitem{wu2021cvt}
H.~Wu, B.~Xiao, N.~Codella, M.~Liu, X.~Dai, L.~Yuan, and L.~Zhang, ``Cvt: Introducing convolutions to vision transformers,'' in {\em Proceedings of the IEEE/CVF international conference on computer vision}, pp.~22--31, 2021.

\bibitem{sandler2018mobilenetv2}
M.~Sandler, A.~Howard, M.~Zhu, A.~Zhmoginov, and L.-C. Chen, ``Mobilenetv2: Inverted residuals and linear bottlenecks,'' in {\em Proceedings of the IEEE conference on computer vision and pattern recognition}, pp.~4510--4520, 2018.

\bibitem{wang2021pyramid}
W.~Wang, E.~Xie, X.~Li, D.-P. Fan, K.~Song, D.~Liang, T.~Lu, P.~Luo, and L.~Shao, ``Pyramid vision transformer: A versatile backbone for dense prediction without convolutions,'' in {\em Proceedings of the IEEE/CVF international conference on computer vision}, pp.~568--578, 2021.

\bibitem{mehta2022separable}
S.~Mehta and M.~Rastegari, ``Separable self-attention for mobile vision transformers,'' {\em arXiv preprint arXiv:2206.02680}, 2022.

\bibitem{maaz2022edgenext}
M.~Maaz, A.~Shaker, H.~Cholakkal, S.~Khan, S.~W. Zamir, R.~M. Anwer, and F.~Shahbaz~Khan, ``Edgenext: efficiently amalgamated cnn-transformer architecture for mobile vision applications,'' in {\em European conference on computer vision}, pp.~3--20, Springer, 2022.

\bibitem{vasu2023fastvit}
P.~K.~A. Vasu, J.~Gabriel, J.~Zhu, O.~Tuzel, and A.~Ranjan, ``Fastvit: A fast hybrid vision transformer using structural reparameterization,'' in {\em Proceedings of the IEEE/CVF International Conference on Computer Vision}, pp.~5785--5795, 2023.

\bibitem{li2023rethinking}
Y.~Li, J.~Hu, Y.~Wen, G.~Evangelidis, K.~Salahi, Y.~Wang, S.~Tulyakov, and J.~Ren, ``Rethinking vision transformers for mobilenet size and speed,'' in {\em Proceedings of the IEEE/CVF International Conference on Computer Vision}, pp.~16889--16900, 2023.

\bibitem{vasu2023mobileone}
P.~K.~A. Vasu, J.~Gabriel, J.~Zhu, O.~Tuzel, and A.~Ranjan, ``Mobileone: An improved one millisecond mobile backbone,'' in {\em Proceedings of the IEEE/CVF conference on computer vision and pattern recognition}, pp.~7907--7917, 2023.

\bibitem{yun2024shvit}
S.~Yun and Y.~Ro, ``Shvit: Single-head vision transformer with memory efficient macro design,'' in {\em Proceedings of the IEEE/CVF Conference on Computer Vision and Pattern Recognition}, pp.~5756--5767, 2024.

\bibitem{yu2022metaformer}
W.~Yu, M.~Luo, P.~Zhou, C.~Si, Y.~Zhou, X.~Wang, J.~Feng, and S.~Yan, ``Metaformer is actually what you need for vision,'' in {\em Proceedings of the IEEE/CVF conference on computer vision and pattern recognition}, pp.~10819--10829, 2022.

\bibitem{russakovsky2015imagenet}
O.~Russakovsky, J.~Deng, H.~Su, J.~Krause, S.~Satheesh, S.~Ma, Z.~Huang, A.~Karpathy, A.~Khosla, M.~Bernstein, {\em et~al.}, ``Imagenet large scale visual recognition challenge,'' {\em International journal of computer vision}, vol.~115, pp.~211--252, 2015.

\bibitem{touvron2021training}
H.~Touvron, M.~Cord, M.~Douze, F.~Massa, A.~Sablayrolles, and H.~J{\'e}gou, ``Training data-efficient image transformers \& distillation through attention,'' in {\em International conference on machine learning}, pp.~10347--10357, PMLR, 2021.

\bibitem{loshchilov2017decoupled}
I.~Loshchilov, ``Decoupled weight decay regularization,'' {\em arXiv preprint arXiv:1711.05101}, 2017.

\bibitem{lin2014microsoft}
T.-Y. Lin, M.~Maire, S.~Belongie, J.~Hays, P.~Perona, D.~Ramanan, P.~Doll{\'a}r, and C.~L. Zitnick, ``Microsoft coco: Common objects in context,'' in {\em Computer Vision--ECCV 2014: 13th European Conference, Zurich, Switzerland, September 6-12, 2014, Proceedings, Part V 13}, pp.~740--755, Springer, 2014.

\bibitem{ross2017focal}
T.-Y. Ross and G.~Doll{\'a}r, ``Focal loss for dense object detection,'' in {\em proceedings of the IEEE conference on computer vision and pattern recognition}, pp.~2980--2988, 2017.

\bibitem{liu2023efficientvit}
X.~Liu, H.~Peng, N.~Zheng, Y.~Yang, H.~Hu, and Y.~Yuan, ``Efficientvit: Memory efficient vision transformer with cascaded group attention,'' in {\em Proceedings of the IEEE/CVF Conference on Computer Vision and Pattern Recognition}, pp.~14420--14430, 2023.

\bibitem{mmdetection}
K.~Chen, J.~Wang, J.~Pang, Y.~Cao, Y.~Xiong, X.~Li, S.~Sun, W.~Feng, Z.~Liu, J.~Xu, Z.~Zhang, D.~Cheng, C.~Zhu, T.~Cheng, Q.~Zhao, B.~Li, X.~Lu, R.~Zhu, Y.~Wu, J.~Dai, J.~Wang, J.~Shi, W.~Ouyang, C.~C. Loy, and D.~Lin, ``{MMDetection}: Open mmlab detection toolbox and benchmark,'' {\em arXiv preprint arXiv:1906.07155}, 2019.

\bibitem{chen2023run}
J.~Chen, S.-h. Kao, H.~He, W.~Zhuo, S.~Wen, C.-H. Lee, and S.-H.~G. Chan, ``Run, don't walk: chasing higher flops for faster neural networks,'' in {\em Proceedings of the IEEE/CVF conference on computer vision and pattern recognition}, pp.~12021--12031, 2023.

\bibitem{howard2019searching}
A.~Howard, M.~Sandler, G.~Chu, L.-C. Chen, B.~Chen, M.~Tan, W.~Wang, Y.~Zhu, R.~Pang, V.~Vasudevan, {\em et~al.}, ``Searching for mobilenetv3,'' in {\em Proceedings of the IEEE/CVF international conference on computer vision}, pp.~1314--1324, 2019.

\end{thebibliography}
\bibliographystyle{ieeetr}
\end{document}